\documentclass[11pt]{article}

\PassOptionsToPackage{hyperfootnotes=false}{hyperref}
\usepackage[final]{acl}

\usepackage{times}
\usepackage{latexsym}

\usepackage[T1]{fontenc}

\usepackage[utf8]{inputenc}

\usepackage{microtype}

\usepackage{inconsolata}

\usepackage{graphicx}

\usepackage{xspace}  
\usepackage{xcolor}  

\usepackage{booktabs}
\usepackage{multirow}
\usepackage{amsmath}
\usepackage{amsfonts}
\usepackage{array}
\usepackage{pifont}
\usepackage{stfloats}

\title{GeoMotionGPT: Geometry-Aligned Motion Understanding with Large Language Models}

\author{
    \textbf{Zhankai Ye}\textsuperscript{1,*} \quad
    \textbf{Bofan Li}\textsuperscript{1,*} \quad
    \textbf{Yukai Jin}\textsuperscript{1} \quad
    \textbf{Shuoqiu Li}\textsuperscript{1}
    \\
    \textbf{Wei Wang}\textsuperscript{2} \quad
    \textbf{Yanfu Zhang}\textsuperscript{3} \quad
    \textbf{Shangqian Gao}\textsuperscript{1,\textdagger} \quad
    \textbf{Xin Liu}\textsuperscript{1,\textdagger}
    \\
    \normalfont\textsuperscript{1}Florida State University \quad
    \textsuperscript{2}Texas Tech University \quad
    \textsuperscript{3}College of William \& Mary
    \\
    \normalfont\small\texttt{\{zy22b, bl21n, yj25b, sl24cs, sg24bi, xliu15\}@fsu.edu} \quad
    \texttt{wei.wang@ttu.edu} \quad
    \texttt{yzhang105@wm.edu}
}

\usepackage{xcolor}
\usepackage{pgf}
\usepackage{subcaption}

\newcommand{\lcell}[1]{%
  \raisebox{-0.18\height}{%
    \parbox{\linewidth}{\centering\small\textbf{#1}}%
  }%
}

\newcommand{\tid}[1]{%
  \begingroup
  \pgfmathparse{mod(#1*0.6180339887,1)}%
  \edef\hue{\pgfmathresult}%
  \textcolor[hsb]{\hue,0.65,0.90}{\ttfamily #1}%
  \endgroup
}

\newcommand{\sysname}{GeoMotionGPT\xspace}

\usepackage{amsthm}

\newcommand{\imgcell}[2]{%
  \raisebox{-0.25\height}{\includegraphics[width=#1]{#2}}%
}

\newcommand{\bluetxt}[1]{\textcolor{blue}{#1}}

\usepackage{pifont}

\begin{document}
\maketitle
\begingroup
\renewcommand{\thefootnote}{}
\footnotetext{\textsuperscript{*}Equal contribution. \quad
\textsuperscript{\textdagger}Corresponding authors.}
\endgroup


\begin{abstract}
Discrete motion tokenization allows large language models (LLMs) to process motion as sequences of vocabulary items, but standard pipelines connect a motion codebook to the LLM only through token IDs. We introduce \sysname, which instead gives the codebook and its corresponding motion-token embeddings a shared orthogonality prior. A decoder-only quantizer uses Gumbel-Softmax assignments and utilization regularization; a fixed sparse isometry transfers the learned codebook Gram matrix to the LLM embedding space at initialization; and soft orthogonality penalties are applied during tokenizer training and LLM fine-tuning. Under the GPT-2 setting, \sysname improves a descriptive aggregate of retrieval and captioning metrics by \textbf{22.4\%} over the strongest reported baseline on HumanML3D and by \textbf{14.4\%} on KIT-ML. Controlled ablations on both datasets attribute these gains to the tokenizer, sparse initialization, and regularization. The results support orthogonality as a useful structural prior while also showing that the model does not dominate every individual metric.
\end{abstract}


\section{Introduction}

Human motion understanding is important for embodied agents that must interpret and act in the physical world~\cite{act,palme}. Recent systems represent motion as discrete tokens and use LLMs for motion understanding, reasoning, and generation~\cite{mmdist,avatargpt,t2mgpt,tm2t}. This interface is attractive because it turns a continuous signal into a sequence that an LLM can consume with its existing language vocabulary.

Most motion-language pipelines train the tokenizer and the language model in separate stages~\cite{t2mgpt,tm2t,motiongpt1,motiongpt2,radarllm}. A vector-quantized model first compresses motion into codebook indices, after which new LLM vocabulary entries are learned for those indices. Token identity links the two stages, but the pairwise relations among the learned code vectors are not transferred explicitly to the initial LLM embeddings. The language model must therefore learn those relations from downstream supervision.

We study whether a simple structural prior can make this handoff more controlled. \sysname\footnote{Project resources: \href{https://github.com/JYe16/GeoMotionGPT}{GitHub} and \href{https://huggingface.co/zy22b/GeoMotionGPT}{Hugging Face}.} encourages the motion code vectors to be approximately orthogonal and uses a fixed sparse isometry to initialize the corresponding LLM motion-token embeddings. The projection preserves the codebook Gram matrix exactly at initialization. During LLM fine-tuning, a soft orthogonality penalty discourages the motion-token embeddings from collapsing into redundant directions while still allowing task-driven adaptation.

Orthogonality is a deliberately limited prior. It does not encode the semantic topology of human motion or align motion vectors directly with text embeddings. Instead, it provides a well-conditioned starting structure for newly introduced tokens without requiring paired contrastive targets. Similar actions can still acquire related contextual representations through the LLM objective.

The framework has three components. First, a decoder-only quantizer uses Gumbel-Softmax assignments~\cite{jang2017categorical}, a straight-through estimator, and utilization regularization so that the codebook can be optimized jointly with reconstruction and geometry objectives. Second, a parameter-free sparse projection places each code dimension in a distinct LLM embedding dimension, preserving all pairwise inner products at initialization. Third, soft orthogonality penalties are applied in both training stages: to the codebook during tokenizer training and to the motion-token embedding block during LLM fine-tuning.

On HumanML3D and KIT-ML, \sysname improves a descriptive aggregate of retrieval and captioning metrics by \textbf{22.4\%} and \textbf{14.4\%}, respectively, over the strongest reported baselines under the GPT-2 setting. The aggregate is not a standard benchmark metric, so we report every component separately and avoid treating it as evidence of uniform dominance. Controlled ablations on both datasets show that removing the tokenizer design, sparse initialization, or orthogonality penalty reduces performance.

Our contributions can be summarized as follows:

\begin{itemize}
    \item We formulate the tokenizer-to-LLM handoff as transfer of a structural prior and use approximate orthogonality to keep newly introduced motion-token directions well conditioned.
    \vspace{-2mm}
    \item We combine a decoder-only Gumbel-Softmax quantizer with a sparse isometric initialization whose preservation of pairwise inner products follows directly from its construction.
    \vspace{-2mm}
    \item Experiments and controlled ablations on HumanML3D and KIT-ML characterize the benefits and trade-offs of the tokenizer, projection, and regularization choices.
\end{itemize}


\section{Related Work}

\noindent{\textbf{Motion Understanding Using LLMs.}}
Motion-language methods differ mainly in how they connect motion and text. Some learn a shared space through contrastive or retrieval objectives~\cite{tmr,motionclip}. Others convert motion into discrete or token-like sequences and apply language-modeling objectives~\cite{tm2t,motiongpt1,motiongpt2,motiongpt3}. Self-supervised motion objectives, including masked modeling, can strengthen these representations~\cite{momask}, while unified formulations cover multiple granularities and interaction settings~\cite{uniframe,mgmotionllm}. More broadly, multimodal language models show that a new modality can be introduced through adapters or projections into an LLM embedding space~\cite{flamingo,blip2}.

\noindent{\textbf{VQ-VAE and Its Variants.}}
Variational autoencoders optimize an evidence lower bound that combines reconstruction with latent regularization~\cite{vae}. Increasing the weight of the KL term can encourage more factorized latent representations, although it may reduce reconstruction fidelity~\cite{betavae}. VQ-VAE instead uses a learned discrete codebook at the bottleneck, with auxiliary losses for codebook learning and encoder commitment~\cite{vqvae}; hierarchical discrete latents extend this design across multiple representation scales~\cite{vavqe2}. Later quantizers simplify the discretization rule, encode residuals with multiple codebooks, improve large-vocabulary utilization, or structure codebook reuse~\cite{fsqvae,rqvae,vqgan100,vqpos}. SoftVQ-VAE explores a differentiable continuous tokenizer for image generation~\cite{chen2025softvq}. Our focus is different: we use differentiable assignments to optimize the geometry and utilization of a motion codebook that will initialize LLM motion-token embeddings.

\begin{figure*}
    \centering
    \resizebox{0.88\textwidth}{!}{%
    \includegraphics[width=1\textwidth]{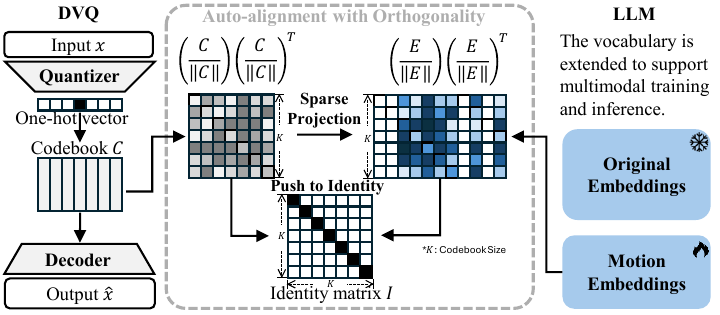}
    }
    \caption{Overview of \sysname. Left: a DVQ motion tokenizer maps an input motion $x$ to discrete codebook indices and reconstructs $\hat{x}$. Middle: orthogonality regularization encourages the normalized codebook Gram matrix to approach the identity, and a fixed sparse isometry preserves this matrix when initializing the motion-token embeddings. Right: the LLM vocabulary is extended with trainable motion-token embeddings while the original text embeddings remain frozen.}
    \label{fig:overall}
    \vspace{-4mm}
\end{figure*}

\noindent{\textbf{Orthogonality in Representation Learning.}}
Orthogonality is a common geometric bias for learning stable and well-conditioned representations.
Spectral control can improve optimization by limiting amplification or suppression of signals and gradients~\cite{svb}.
Orthogonal transformations also preserve inner products and norms, a property used in work on compatible and approximately isometric representations~\cite{lambdaorth,isoortho,ole}. Practical approaches include Stiefel-manifold optimization, normalization methods, and partial orthogonalization~\cite{own,odnn}. Related constraints have been studied in convolutional networks, initialization methods, and attention models~\cite{ocnn,orandinit,otrans,ovit}.

\noindent{\textbf{Positioning of \sysname.}}
Prior motion tokenizers primarily optimize reconstruction and code usage~\cite{t2mgpt,tm2t,momask,hong2025egolm}. DVQ adds differentiable assignments so that an explicit codebook geometry penalty can be optimized together with those objectives. Prior orthogonality methods usually regularize a single representation space~\cite{svb,own,odnn,ocnn,otrans,ovit}. We apply the same form of regularizer to the Stage-1 codebook and the Stage-2 motion-token embeddings, with a fixed isometry transferring the codebook Gram matrix at initialization. This is a structural bridge between two training stages, not a direct semantic alignment between motion and text.


\section{Our Approach}
\subsection{Orthogonality as a Shared Structural Prior}
\label{sec:alignment}

Let $\mathcal{M}$ denote the continuous motion space, $\mathcal{Z}=\{1,\dots,K\}$ the discrete motion-token index set, and $\mathcal{S}$ the LLM embedding space.

Existing approaches typically learn a quantization map $q:\mathcal{M}\rightarrow\mathcal{Z}$ and a separate embedding map $\phi:\mathcal{Z}\rightarrow\mathcal{S}$. The index $k\in\mathcal{Z}$ links the two stages, but the pairwise inner products among code vectors are not normally used to initialize the corresponding LLM embeddings.

We use orthogonality as a shared structural prior for this handoff. As illustrated in Figure~\ref{fig:overall}, the motion codebook $\mathbf{C} = \{\mathbf{c}_1, \dots, \mathbf{c}_K\} \subset \mathbb{R}^D$ is encouraged to approximate an orthonormal set:
\begin{equation}
    \langle \mathbf{c}_i, \mathbf{c}_j \rangle \approx \delta_{ij},
    \label{eq:ortho_basis} 
\end{equation}
where $\delta_{ij}$ is the Kronecker delta. Exact row orthogonality is possible only when $K\leq D$; our experiments use $K=D=512$, and training imposes a soft penalty rather than a hard constraint.

To enforce this condition during training, we apply orthogonal regularization to the codebook. Let $\hat{\mathbf{C}}$ be the row-normalized codebook where $\hat{\mathbf{c}}_k = \mathbf{c}_k / \|\mathbf{c}_k\|_2$. We compute the Gram matrix $\mathbf{G} = \hat{\mathbf{C}} \hat{\mathbf{C}}^\top$ and define the orthogonal loss as:
\begin{equation}
    \mathcal{L}_{\mathrm{ortho}} = \left\| \mathbf{G} - \mathbf{I}_K \right\|_F^2.
    \label{eq:loss_ortho}
\end{equation}
This loss discourages redundant code directions.

Two properties of this objective matter.
First, it is supervision-free and closed form: unlike contrastive or metric-based alignment, it needs no positive and negative pairs, similarity targets, or text encoder because the target is the identity matrix.
Second, it is a \textit{soft} penalty rather than a hard constraint, and not an assertion that similar actions must be mutually unrelated.
Eq.~\ref{eq:loss_ortho} is one term in a composite objective dominated by reconstruction during tokenizer training and by the task loss during fine-tuning (Section~\ref{sec:regulation}).
Its role is to reduce directional redundancy. The objective does not preserve the neighborhood topology of the original motion space, and semantic relations among actions must still be learned from the task objective.

\subsection{Structure-Preserving Sparse Projection}
\label{sec:projection}

We next transfer the learned codebook Gram matrix to the LLM motion-token embeddings at initialization. A learned dense map has no built-in inner-product preservation guarantee, so we instead use a fixed sparse projection.

We map the $D$-dimensional motion codes into the higher-dimensional LLM space $\mathbb{R}^{D'}$ ($D' \gg D$) by distributing them across randomly selected active dimensions.
Specifically, we define a fixed projection matrix $\mathbf{P} \in \{0,1\}^{D' \times D}$ initialized by randomly selecting $D$ unique row indices $\mathcal{I} \subset \{1, \dots, D'\}$ to act as identity mappings, while setting all other entries to zero.
As illustrated in Figure~\ref{fig:overall}, the embedding is computed as:

\vspace{-4mm}
\begin{equation}
    \mathbf{e}_k = \mathbf{P}\mathbf{c}_k.
\end{equation}
\vspace{-6mm}

\noindent Intuitively, this operation scatters the motion code values into the high-dimensional vector $\mathbf{e}_k$ at random positions, filling the remaining $D' - D$ dimensions with zeros.

This projection is an isometric embedding and preserves pairwise inner products regardless of which unique row indices are selected.

\noindent\textbf{Lemma.}
\textit{Let $\mathbf{P}$ be a sparse projection matrix where each column contains exactly one '1' at a unique row index and '0' elsewhere. If the original codebook vectors $\{\mathbf{c}_k\}$ are pairwise orthogonal, then the projected embeddings $\{\mathbf{e}_k\}$ are also pairwise orthogonal.}

\noindent\textbf{Proof.}
The inner product in the LLM space is:

\vspace{-6mm}
\begin{equation*}
    \mathbf{e}_i^\top \mathbf{e}_j
    =
    (\mathbf{P}\mathbf{c}_i)^\top (\mathbf{P}\mathbf{c}_j)
    =
    \mathbf{c}_i^\top (\mathbf{P}^\top \mathbf{P}) \mathbf{c}_j.
\end{equation*}
\vspace{-6mm}

\noindent Since $\mathbf{P}$ maps source dimension to a unique target dimension without overlap, the columns of $\mathbf{P}$ are orthonormal. Thus, $\mathbf{P}^\top \mathbf{P} = \mathbf{I}_D$. It follows that:

\vspace{-3mm}
\begin{equation*}
    \mathbf{e}_i^\top \mathbf{e}_j
    =
    \mathbf{c}_i^\top \mathbf{I}_D \mathbf{c}_j
    =
    \mathbf{c}_i^\top \mathbf{c}_j.
\end{equation*}
\vspace{-6mm}

\noindent If $\mathbf{c}_i \perp \mathbf{c}_j$, then $\mathbf{e}_i \perp \mathbf{e}_j$.

Two clarifications define the scope of this result.
\textit{(i) The guarantee also covers approximate orthogonality.}
Since $\mathbf{P}^\top\mathbf{P}=\mathbf{I}_D$ by construction, the argument applies to \emph{every} pair of codes. Writing $\hat{\mathbf{E}}$ for the row-normalized projected embeddings, $\hat{\mathbf{E}}\hat{\mathbf{E}}^\top = \hat{\mathbf{C}}\hat{\mathbf{C}}^\top$, hence $\|\hat{\mathbf{E}}\hat{\mathbf{E}}^\top-\mathbf{I}_K\|_F = \|\hat{\mathbf{C}}\hat{\mathbf{C}}^\top-\mathbf{I}_K\|_F$: the projection transfers the \textit{exact degree} of approximate orthogonality attained by the soft penalty, which is the regime that actually arises in training.
\textit{(ii) It is an initialization mechanism, not a permanent constraint.}
$\mathbf{P}$ is fixed, but the motion-token embeddings are optimized in Stage~2. The equality of the two Gram matrices therefore holds at initialization only. Stage-2 regularization encourages approximate orthogonality after fine-tuning, but it does not preserve exact codebook-to-embedding correspondence.

\vspace{-1mm}
\subsection{Decoder-Only Vector Quantization (DVQ)}
\label{sec:dvq}

Because the sparse projection preserves the codebook Gram matrix at initialization, the first stage must learn a useful codebook structure.
Standard VQ-VAE uses a non-differentiable nearest-neighbor assignment and relies on surrogate updates for the encoder and codebook. We instead use a Gumbel-Softmax relaxation so that reconstruction, utilization, and geometry terms can be optimized in one computation graph.

Given motion input $x$, the quantizer produces logits $\mathbf{z}=Q(x)\in\mathbb{R}^{K}$. We sample a relaxed assignment with Gumbel-Softmax~\cite{jang2017categorical} and discretize it for the forward pass:

\vspace{-3mm}
\begin{equation}
\label{eq:gumbelsoftmax}
\begin{aligned}
\mathbf{y}_{\text{soft}} &= \mathrm{GumbelSoftmax}(\mathbf{z}; \tau), \\
\mathbf{y}_{\text{hard}} &= \mathrm{one\text{-}hot}\Big(\arg\max_{k} \; y_{\text{soft},k}\Big),
\end{aligned}
\end{equation}
\vspace{-3mm}


\noindent where $\tau$ is the temperature. The selected motion embedding is

\vspace{-3mm}
\begin{equation}
    {\mathbf{h}} = \mathbf{y}_{\text{hard}}^\top \mathbf{C}.
\end{equation}
\vspace{-3mm}

\noindent where $y_{\text{soft},k}$ is the $k$-th entry of $\mathbf{y}_{\text{soft}}$. We use the straight-through estimator (STE)~\cite{bengio2013estimating}: the forward pass uses $\mathbf{y}_{\text{hard}}$, while the backward pass follows the relaxed assignment. The decoder output is $\hat{x} = D(\mathbf{h})$.

The STE gradient is biased, but its forward path matches the discrete representation used at inference. Without the STE, training optimizes only the continuous relaxation; in our ablation this mismatch destabilizes late training and leaves 323 of 512 codes unused (Table~\ref{tab:diag}). Gumbel noise is used only during training. At inference, tokenization is a deterministic $\arg\max$ over the logits.

We also regularize {codebook utilization} to prevent token collapse.
Let $q_k$ be the normalized mini-batch assignment frequency of code $k$, computed from the scheduled soft-to-hard utilization statistics described in Appendix~\ref{sec:appendix_dvq}.
To enforce a balanced distribution, we maximize its self-entropy:
\vspace{-3mm}
\begin{equation}\label{eq:loss_util}
\mathcal{L}_{\mathrm{util}}
=
- H(\mathbf{q})
=
\sum_{k=1}^{K} q_k \log q_k,
\end{equation}
\vspace{-3mm}

\noindent with the convention $0\log0=0$. Minimizing this term favors higher-entropy assignment frequencies. It encourages balanced use but does not guarantee a uniform distribution.
The two regularizers are complementary: $\mathcal{L}_{\mathrm{util}}$ shapes the \textit{marginal} distribution over codes (how often each is selected), whereas $\mathcal{L}_{\mathrm{ortho}}$ shapes their \textit{geometry} (how they are arranged). Balanced usage alone does not stop codes from becoming near-duplicates, and orthogonality alone does not stop a few codes from dominating assignments.

\vspace{-1mm}
\subsection{Two-Stage Orthogonality Regularization}
\label{sec:regulation}

We apply the same orthogonality prior in two stages, as illustrated in Figure~\ref{fig:overall}.

In Stage~1, we train DVQ with a composite objective that adds geometry and utilization terms to the reconstruction loss $\mathcal{L}_{\mathrm{rec}} = \|x-\hat{x}\|_2^2$:

\vspace{-6mm}
\begin{equation}
    \mathcal{L}_{\mathrm{DVQ}} = \mathcal{L}_{\mathrm{rec.}} + \lambda_{\mathrm{ortho}} \mathcal{L}_{\mathrm{ortho}} + \lambda_{\mathrm{util}} \mathcal{L}_{\mathrm{util}}.
\end{equation}
\vspace{-6mm}

\noindent $\mathcal{L}_{\mathrm{ortho}}$ and $\mathcal{L}_{\mathrm{util}}$ are defined in Eq.~\ref{eq:loss_ortho} and Eq.~\ref{eq:loss_util}.

In Stage~2, we initialize the LLM motion-token block $\mathbf{E}\in\mathbb{R}^{K\times D'}$ by sparse projection and extend the original token embedding matrix $\mathbf{E}_{\mathrm{org}}\in\mathbb{R}^{N\times D'}$ as $\mathbf{E}_{\mathrm{new}}=[\mathbf{E}_{\mathrm{org}};\mathbf{E}]\in\mathbb{R}^{(N+K)\times D'}$. We freeze the original text embeddings and optimize the motion-token embeddings together with the trainable LLM parameters. A soft orthogonality term continues to discourage redundancy within the motion-token block:

\vspace{-6mm}
\begin{equation}
\label{eq:tuning}
\begin{aligned}
\mathcal{L}_{\mathrm{tuning}}
&=
\underbrace{
-
\mathbb{E}_{(X,Y)}
\sum_{t=1}^{|Y|}
\log p_{\theta}\!\left(
y_t \mid y_{<t}, X
\right)
}_{\text{task loss}} \\
&\quad
+
\underbrace{
\lambda_{\mathrm{orth}}'
\left\|
\hat{\mathbf{E}}\hat{\mathbf{E}}^{\top}
-
\mathbf{I}_K
\right\|_F^2
}_{\text{orthogonality regularization}},
\end{aligned}
\end{equation}
\vspace{-3mm}

\noindent where $\hat{\mathbf{E}}\in\mathbb{R}^{K\times D'}$ contains row-normalized motion embeddings, $\hat{\mathbf{e}}_k=\mathbf{e}_k/\|\mathbf{e}_k\|_2$. The input $X=[X_{\text{prompt}},X_{\text{motion}}]$ contains prompt and DVQ motion tokens, $Y$ is the target response, and $p_{\theta}$ is the LLM.

\paragraph{Why two stages, and what bridges them.}
Two-stage training is common in motion-language systems: a tokenizer is learned first, then an LLM is adapted. Our bridge carries more than token identity. Stage~1 shapes the codebook Gram matrix, and the sparse projection copies that matrix to the initial motion-token embeddings. The Stage-2 penalty then encourages low pairwise cosine similarity as the embeddings adapt. It does not constrain each tuned embedding to remain close to its projected Stage-1 code. Contrastive learning could provide semantic supervision earlier, but it would require paired text representations during tokenizer training; our tokenizer remains motion-only.



\vspace{-1mm}
\section{Experiment}

We describe the experimental setup, compare with prior methods, and report controlled ablations and diagnostics.


\vspace{-1mm}
\subsection{Experimental Settings}
\label{sec:experimental_settings}
We evaluate three language backbones: GPT-2~\cite{gpt2}, Qwen3 0.6B~\cite{qwen3}, and Llama 3.2 1B~\cite{llama3.2}.
GPT-2 is fully fine-tuned, while Qwen3 0.6B and Llama 3.2 1B are adapted with LoRA~\cite{hu2022lora}.

All models are trained and evaluated on two standard benchmarks: HumanML3D~\cite{humanml3d} and KIT-ML~\cite{plappert2016kit}.
HumanML3D is a benchmark for text-conditioned human motion understanding and generation. We follow its preprocessing pipeline and represent each frame as a 263-dimensional feature vector; each motion has three text captions.
KIT-ML is a widely used benchmark with more diverse action descriptions, which provides a complementary testbed for cross-dataset evaluation.

We follow the motion-to-text evaluation protocol used by MotionGPT3~\cite{motiongpt3}.

\paragraph{Training protocol.}
DVQ uses a 512-code, 512-dimensional codebook and is trained for 500 epochs with batch size 512. We use AdamW with a base learning rate of $2\times10^{-4}$, 3\% linear warmup, and cosine decay; quantizer parameters use a learning rate of $10^{-4}$. The Gumbel-Softmax temperature stays at $0.4$ for 300 epochs, decays exponentially to $0.01$ over the next 100, and then remains fixed. The utilization statistic uses soft assignments for 150 epochs; its hardness coefficient then increases linearly from 0 to 1 over 50 epochs. Training uses stochastic Gumbel-Softmax assignments, whereas inference applies deterministic $\arg\max$ tokenization. For Stage-2 motion understanding, we train for 100 epochs with batch size 320, learning rate $10^{-4}$, weight decay $10^{-2}$, and cosine scheduling. Unless otherwise specified, each experiment runs on one NVIDIA B200 GPU. Appendix~\ref{sec:appendix_dvq} gives the remaining optimizer and adaptation details.


\begin{figure}[t]
    \centering
    \begin{subfigure}[t]{0.49\linewidth}
        \centering
        \includegraphics[width=\linewidth,trim=4 4 4 2,clip]{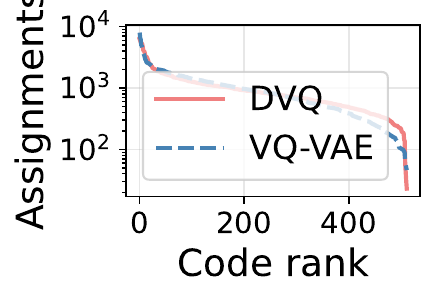}
        \caption{Ranked usage}
        \label{fig:vq-usage-curve}
    \end{subfigure}
    \hfill
    \begin{subfigure}[t]{0.49\linewidth}
        \centering
        \includegraphics[width=\linewidth,trim=4 4 4 2,clip]{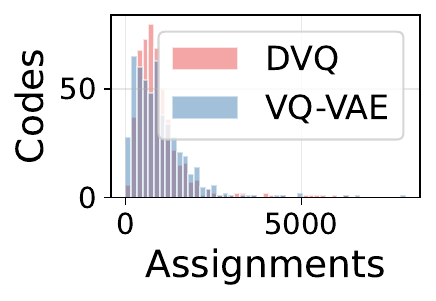}
        \caption{Usage histogram}
        \label{fig:vq-usage-count}
    \end{subfigure}
    \vspace{-2mm}
    \caption{Codebook usage on the evaluation set. Both tokenizers use all 512 codes, but DVQ has a less skewed count distribution than the VQ-VAE baseline.}
    \vspace{-6mm}
\end{figure}

\begin{table*}[t]
    \centering
    \small
    \setlength{\tabcolsep}{2.5pt}
    \caption{Motion-to-text results on HumanML3D under the GPT-2 setting. \sysname has the highest descriptive aggregate, 22.4\% above MotionGPT3, but does not lead on every retrieval metric.}
    \renewcommand{\arraystretch}{1}
    \begin{tabular}{lcccccccccc}
    \toprule
    Approach & R@1 & R@2 & R@3 & MMDist$\downarrow$ & BLEU-1$\uparrow$ & BLEU-4$\uparrow$ & R-L$\uparrow$ & CIDEr$\uparrow$ & BERTScore$\uparrow$ & Avg.$\uparrow$ \\
    \midrule
    \textbf{Real} & 0.523 & 0.725 & 0.828 & 2.901 & - & - & - & - & - & - \\
    \midrule
    TM2T~\cite{tm2t}  & 0.516 & - & 0.823 & 2.935 & 48.90 & 7.00 & 38.10 & 16.80 & 32.20 & 34.99 \\
    MotionGPT~\cite{motiongpt1}  & 0.543 & - & 0.827 & 2.821 & 48.20 & 12.50 & 37.40 & 29.20 & 32.40 & 38.03 \\
    LaMPM2T~\cite{lampm2t}  & 0.547 & - & 0.831 & 2.808 & 47.80 & 13.04 & 37.10 & 28.90 & 32.70 & 38.07\\
    MoTe~\cite{mote} & \textbf{0.577} & - & \textbf{0.871} & 2.649 & 46.70 & 11.15 & 37.40 & 31.50 & 30.30 & 38.24 \\
    MotionGPT3~\cite{motiongpt3} & 0.573 & \textbf{0.773} & 0.864 & \textbf{2.430} & 59.08 & 19.41 & 46.18 & 28.72 & 35.23 & 43.71 \\
    \midrule
    \sysname (Ours) & 0.533 & 0.729 & 0.817 & 2.680 & \textbf{65.65} & \textbf{25.88} & \textbf{51.32} & \textbf{59.71} & \textbf{49.03} & \textbf{53.48} \\
    \bottomrule
    \end{tabular}
    \label{tab:main}
    \vspace{-1mm}
\end{table*}

\subsection{Evaluation Metrics}

To assess codebook quality, we compare the per-code usage distributions of \sysname and a VQ-VAE baseline over the evaluation set, reporting codebook utilization (the percentage of codes with non-zero usage) and the standard deviation of usage counts, which quantifies how concentrated the assignments are.
For a compact descriptive summary, we report an aggregate that combines retrieval and captioning metrics. It is not a standard benchmark measure and is not used to replace the individual metrics. For method $m$, let $\mathcal{K}_m\subseteq\{1,2,3\}$ contain the reported R-Precision cutoffs and define $\bar{R}_m=|\mathcal{K}_m|^{-1}\sum_{k\in\mathcal{K}_m}R_k$. The aggregate is

\vspace{-3mm}
\begin{equation*}
\mathrm{Avg}
= \frac{
100\cdot\bar{R}_m
+ B_1 + B_4 + \mathrm{R\!-\!L} + C + S
}{6},
\label{eq:avg}
\end{equation*}
\vspace{-6mm}

\noindent where $B_1$ and $B_4$ denote BLEU-1 and BLEU-4~\cite{bleu}, $\mathrm{R\!-\!L}$ denotes ROUGE-L~\cite{rouge}, $C$ is CIDEr~\cite{cider}, and $S$ is BERTScore~\cite{bertscore}. R-Precision~\cite{rprecision} is scaled by 100. Some published baselines omit R@2, so their $\bar{R}_m$ averages R@1 and R@3; rows with all three recalls average all three. This difference in metric availability is another reason to treat the aggregate as descriptive only.

\subsection{Codebook Distribution Analysis}
\label{sec:codebook_analysis}

\begin{table*}[t]
    \centering
    \small
    \setlength{\tabcolsep}{2.5pt}
    \caption{Motion-to-text results on KIT-ML~\cite{plappert2016kit} under the GPT-2 setting. \sysname has the highest descriptive aggregate, 14.4\% above MotionGPT.}
    \vspace{-2mm}
    \renewcommand{\arraystretch}{1}
    \begin{tabular}{lcccccccccc}
    \toprule
    Approach & R@1 & R@2 & R@3 & MMDist$\downarrow$ & BLEU-1$\uparrow$ & BLEU-4$\uparrow$ & R-L$\uparrow$ & CIDEr$\uparrow$ & BERTScore$\uparrow$ & Avg.$\uparrow$ \\
    \midrule
    TM2T~\cite{tm2t} & \textbf{0.328} & 0.504 & 0.617 & 5.129 & 36.38 & 10.11 & \textbf{49.11} & 38.82 & 24.75 & 34.58 \\
    MotionGPT~\cite{motiongpt1} & 0.148 & 0.270 & 0.348 & 7.107 & 44.04 & \textbf{19.45} & 44.76 & 47.46 & \textbf{37.56} & 36.46 \\
    MotionGPT3~\cite{motiongpt3} & 0.201 & 0.333 & 0.458 & 5.706 & \textbf{44.55} & 17.48 & 45.21 & 16.73 & 26.72 & 30.63 \\
    \midrule
    \sysname (Ours) & \textbf{0.328} & \textbf{0.574} & \textbf{0.676} & \textbf{3.565} & 44.11 & 15.69 & 42.48 & \textbf{58.81} & 36.57 & \textbf{41.71} \\
    \bottomrule
    \end{tabular}
    \label{tab:kit_ml}
    \vspace{-3mm}
\end{table*}

\begin{table}[t]
    \centering
    \small
    \setlength{\tabcolsep}{3pt}
    \caption{Numerical codebook statistics under an identical codebook size ($K{=}512$). Lower Max Count, Std., CV, and Gini all indicate a less skewed usage distribution; Rec.\ is the final validation reconstruction loss.}
    \vspace{-2mm}
    \renewcommand{\arraystretch}{1}
    \resizebox{\linewidth}{!}{
    \begin{tabular}{lccccccc}
    \toprule
    Tokenizer & Used & Dead & Max Cnt.$\downarrow$ & Std.$\downarrow$ & CV$\downarrow$ & Gini$\downarrow$ & Rec.$\downarrow$ \\
    \midrule
    VQ-VAE & 512 & 0 & 7,971 & 840.36 & 0.865 & 0.405 & \textbf{0.2421} \\
    DVQ (Ours) & 512 & 0 & \textbf{6,300} & \textbf{791.20} & \textbf{0.814} & \textbf{0.358} & 0.3597 \\
    \bottomrule
    \end{tabular}
    }
    \label{tab:codebook_stats}
    \vspace{-6mm}
\end{table}

We first examine how often each learned code is selected on the evaluation set.
As shown in Fig.~\ref{fig:vq-usage-curve}, both methods exhibit a heavy-tailed usage pattern where a small subset of codes is used frequently, while many codes are used less often.
Compared with the VQ-VAE baseline, DVQ is less skewed: the most frequent codes are less dominant, and more codes occupy moderate usage ranges.
The histogram in Fig.~\ref{fig:vq-usage-count} shows the same pattern. Because both methods use every code at least once, these plots measure balance rather than the number of dead codes.

\paragraph{Numerical codebook statistics.}
Table~\ref{tab:codebook_stats} quantifies this trend.
Both tokenizers keep all 512 codes alive, so utilization alone does not separate them; the dispersion statistics do.
DVQ lowers the most frequent code count from 7{,}971 to 6{,}300 and reduces every measure of skew: usage standard deviation, coefficient of variation, and Gini coefficient.
For transparency we also report reconstruction quality, where VQ-VAE is in fact stronger (final validation reconstruction loss 0.2421 vs.\ 0.3597).
DVQ's role in our framework is therefore not better raw reconstruction. Its differentiable assignment allows the codebook geometry penalty to be optimized together with reconstruction, which is evaluated in Section~\ref{sec:ablations}.




\begin{table*}[t]
    \centering
    \small
    \setlength{\tabcolsep}{2.2pt}
    \caption{Ablation on HumanML3D under GPT-2. \textit{Baseline} removes geometry-preserving initialization and orthogonality regularization. We vary the tokenizer, projection, regularizer, and regularization weight. The corresponding KIT-ML ablation appears in Table~\ref{tab:kit_ablation}.}
    \renewcommand{\arraystretch}{1}
    \resizebox{\linewidth}{!}{
    \begin{tabular}{llcccccccccc}
    \toprule
    Study & Setting & R@1 & R@2 & R@3 & MMDist$\downarrow$ & BLEU-1$\uparrow$ & BLEU-4$\uparrow$ & R-L$\uparrow$ & CIDEr$\uparrow$ & BERTScore$\uparrow$ & Avg.$\uparrow$ \\
    \midrule
    Baseline & Init \ding{55}, R=0 & 0.488 & 0.668 & 0.712 & 2.943 & 58.70 & 20.41 & 46.52 & 44.80 & 41.90 & 45.77 \\
    \midrule
    Tokenizer & PQ-VAE & 0.534 & 0.716 & 0.804 & 2.920 & 60.03 & 20.97 & 46.35 & 44.06 & 41.14 & 46.84 \\
    Tokenizer & DVQ (Ours) & 0.533 & 0.729 & 0.817 & 2.680 & 65.65 & 25.88 & 51.32 & 59.71 & 49.03 & 53.48 \\
    \midrule
    Projection & Linear & 0.486 & 0.665 & 0.754 & 4.200 & 54.34 & 16.73 & 40.32 & 33.54 & 30.19 & 39.77 \\
    Projection & Sparse (Ours) & 0.533 & 0.729 & 0.817 & 2.680 & 65.65 & 25.88 & 51.32 & 59.71 & 49.03 & 53.48 \\
    \midrule
    Regularization & Pairwise Cosine Dist. & 0.513 & 0.691 & 0.781 & 3.070 & 59.34 & 20.75 & 45.73 & 43.71 & 41.53 & 46.20 \\
    Regularization & Orthogonal (Ours) & 0.533 & 0.729 & 0.817 & 2.680 & 65.65 & 25.88 & 51.32 & 59.71 & 49.03 & 53.48 \\
    \midrule
    Ortho. Weight & Init \ding{51}, $R=0$ & 0.525 & 0.721 & 0.812 & 2.820 & 57.30 & 19.90 & 47.40 & 47.40 & 42.90 & 47.23 \\
    Ortho. Weight & Init \ding{51}, $R=10^{-4}$ & 0.530 & 0.727 & 0.819 & 2.640 & 63.02 & 24.41 & 49.14 & 55.47 & 44.54 & 50.96 \\
    Ortho. Weight & Init \ding{51}, $R=10^{-3}$ & 0.530 & 0.727 & 0.819 & 2.650 & 63.02 & 22.45 & 49.13 & 55.47 & 44.54 & 50.96 \\
    Ortho. Weight & Init \ding{51}, $R=10^{-2}$ & 0.533 & 0.729 & 0.817 & 2.680 & 65.65 & 25.88 & 51.32 & 59.71 & 49.03 & 53.48 \\
    Ortho. Weight & Init \ding{51}, $R=10^{-1}$ & 0.540 & 0.742 & 0.831 & 2.670 & 60.11 & 21.94 & 49.04 & 51.87 & 43.63 & 49.50 \\
    Ortho. Weight & Init \ding{51}, $R=1$ & 0.530 & 0.727 & 0.819 & 2.650 & 63.02 & 24.41 & 49.17 & 55.47 & 44.54 & 50.97 \\
    Ortho. Weight & Init \ding{55}, $R=10^{-2}$ & 0.509 & 0.699 & 0.804 & 2.800 & 63.66 & 24.29 & 47.14 & 50.69 & 41.49 & 49.06 \\
    \bottomrule
    \end{tabular}
    }
    \label{tab:ablation}
    \vspace{-2mm}
\end{table*}

\subsection{Performance on Motion Understanding}
\label{sec:motion_understanding}

Table~\ref{tab:main} compares \sysname with published HumanML3D results under the GPT-2 setting.
Relative to MotionGPT3, the descriptive Average increases by \textbf{22.4\%}. The largest differences are in caption metrics: CIDEr increases by \textbf{107.9\%}, BLEU-4 by \textbf{33.3\%}, and BERTScore by \textbf{39.2\%}~\cite{motiongpt3}.
Retrieval recall is slightly below the best method, and MMDist is slightly higher.

On KIT-ML~\cite{plappert2016kit} (Table~\ref{tab:kit_ml}), \sysname has the highest Average (41.71), \textbf{14.4\%} above MotionGPT. It also has the best R@2, R@3, MMDist, and CIDEr in the table. MotionGPT remains better on BLEU-4 and BERTScore, and TM2T ties our R@1 while leading on ROUGE-L.

\paragraph{Retrieval and captioning trade-off.}
The two metric families move differently across datasets. On HumanML3D our recalls sit marginally below MoTe and MotionGPT3 while captioning improves by a wide margin; on KIT-ML the pattern partly inverts.
We therefore do not claim that \sysname dominates every metric against every baseline, and cross-method comparisons of individual metrics are in any case only partially informative, since different designs respond differently to dataset characteristics and emphasize retrieval-style alignment versus generation quality.
The controlled ablations within one framework provide cleaner evidence than the cross-paper aggregate. Replacing any core component reduces the aggregate on both datasets, although individual metrics do not always move together (Section~\ref{sec:ablations}).

\subsection{Transfer across LLM Backbones}
\label{sec:backbone_transfer}

\begin{table*}[t]
    \centering
    \small
    \setlength{\tabcolsep}{3pt}
    \caption{Backbone transfer on HumanML3D using the adaptation strategy specified for each language backbone.}
    \vspace{-2mm}
    \renewcommand{\arraystretch}{1}
    \resizebox{\linewidth}{!}{
    \begin{tabular}{lccccccccccc}
    \toprule
    Backbone & Adapt. & R@1 & R@2 & R@3 & MMDist$\downarrow$ & BLEU-1$\uparrow$ & BLEU-4$\uparrow$ & R-L$\uparrow$ & CIDEr$\uparrow$ & BERTScore$\uparrow$ & Avg.$\uparrow$ \\
    \midrule
    GPT-2 & Full & 0.533 & 0.729 & 0.817 & 2.680 & 65.65 & 25.88 & 51.32 & 59.71 & 49.03 & 53.48 \\
    Qwen3 0.6B & LoRA & 0.376 & 0.550 & 0.656 & 3.980 & 62.06 & 22.45 & 47.65 & 45.98 & 42.43 & 45.55 \\
    Llama 3.2 1B & LoRA & 0.444 & 0.644 & 0.740 & 3.220 & 63.80 & 24.18 & 49.32 & 50.60 & 45.45 & 49.05 \\
    \bottomrule
    \end{tabular}
    }

    \label{tab:backbone_transfer}
    \vspace{-3mm}
\end{table*}

\begin{table}[t]
    \ContinuedFloat
    \centering
    \small
    \setlength{\tabcolsep}{2.6pt}
    \caption{Controlled Qwen3 0.6B comparison on HumanML3D with identical LoRA adaptation. \textit{Baseline} removes all proposed components.}
    \vspace{-2mm}
    \renewcommand{\arraystretch}{1}
    \resizebox{\linewidth}{!}{
    \begin{tabular}{lcccccc}
    \toprule
    Setting & R@1 & R@3 & MMDist$\downarrow$ & BLEU-4$\uparrow$ & CIDEr$\uparrow$ & Avg.$\uparrow$ \\
    \midrule
    Baseline & 0.348 & 0.588 & 4.490 & 10.81 & 21.02 & 29.96 \\
    \sysname & \textbf{0.376} & \textbf{0.656} & \textbf{3.980} & \textbf{22.45} & \textbf{45.98} & \textbf{45.55} \\
    \bottomrule
    \end{tabular}
    }
    \label{tab:qwen_controlled}
    \vspace{-3mm}
\end{table}

Table~\ref{tab:backbone_transfer} identifies GPT-2 with full fine-tuning as the strongest tested configuration, but the comparison is not controlled: the two larger backbones use LoRA because of our compute budget. Its single-column Qwen3 panel holds the backbone and LoRA settings fixed and varies only our components. Under this matched setup, \sysname raises Average from 29.96 to 45.55, a \textbf{52.0\%} relative increase. Retrieval improves (R@1 $0.348\rightarrow0.376$; MMDist $4.490\rightarrow3.980$), and captioning improves as well (CIDEr $21.02\rightarrow45.98$). This controlled result supports transfer beyond GPT-2 under LoRA, but it does not establish how the backbones would rank under identical full fine-tuning.


\subsection{Ablation Studies}
\label{sec:ablations}


\begin{figure}[t]
    \vspace{-2mm}
    \centering
    \includegraphics[width=\linewidth]{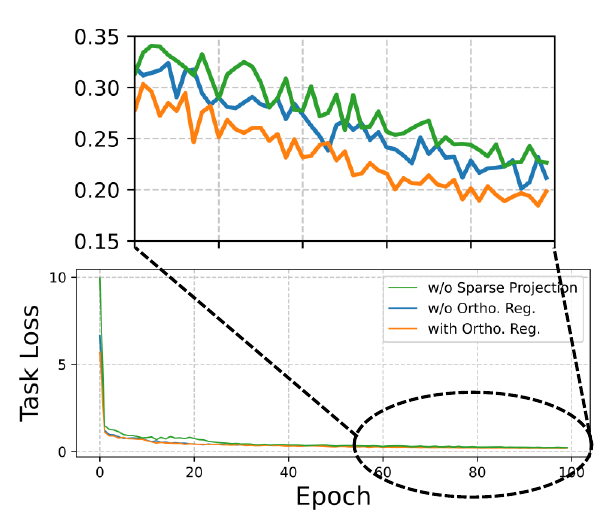}
    \vspace{-6mm}
    \caption{Stage-2 training loss with the full method, without orthogonality loss, and without sparse-projection initialization.}
    \vspace{-4mm}
    \label{fig:total-loss-compare}
\end{figure}

\begin{table*}[t]
    \centering
    \caption{Representative motion-to-text outputs. Blue text marks details recovered by \sysname. The first output also adds an unsupported picking action, and the third omits the rightward part of the reference.}

    \setlength{\tabcolsep}{4pt}
    \renewcommand{\arraystretch}{1.0}

    \begin{tabular}{@{} m{0.125\textwidth} m{0.195\textwidth} m{0.195\textwidth} m{0.195\textwidth} m{0.195\textwidth} @{}}
        \toprule
        \lcell{GT Motions}
        &
        \imgcell{\linewidth}{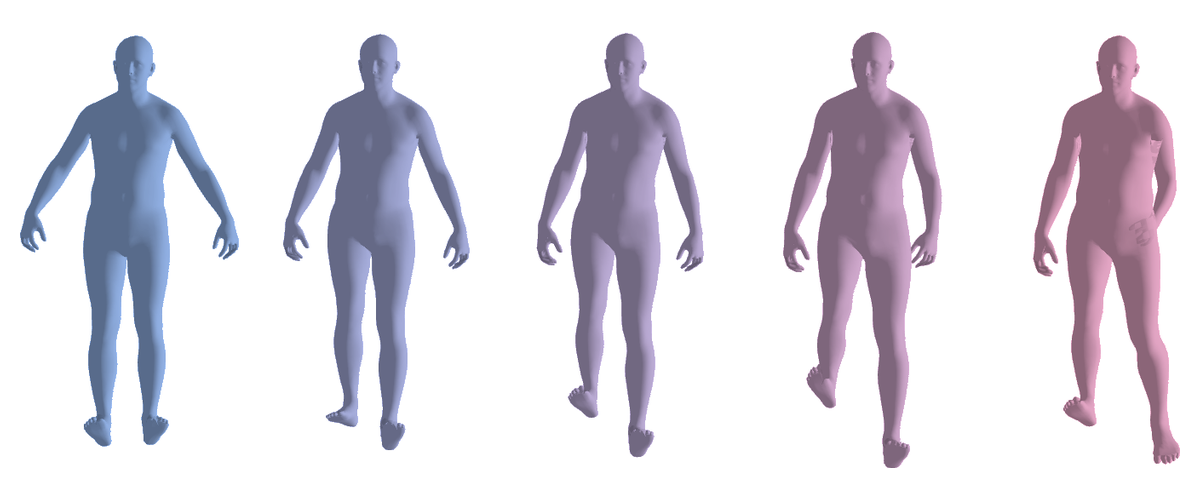}
        &
        \imgcell{\linewidth}{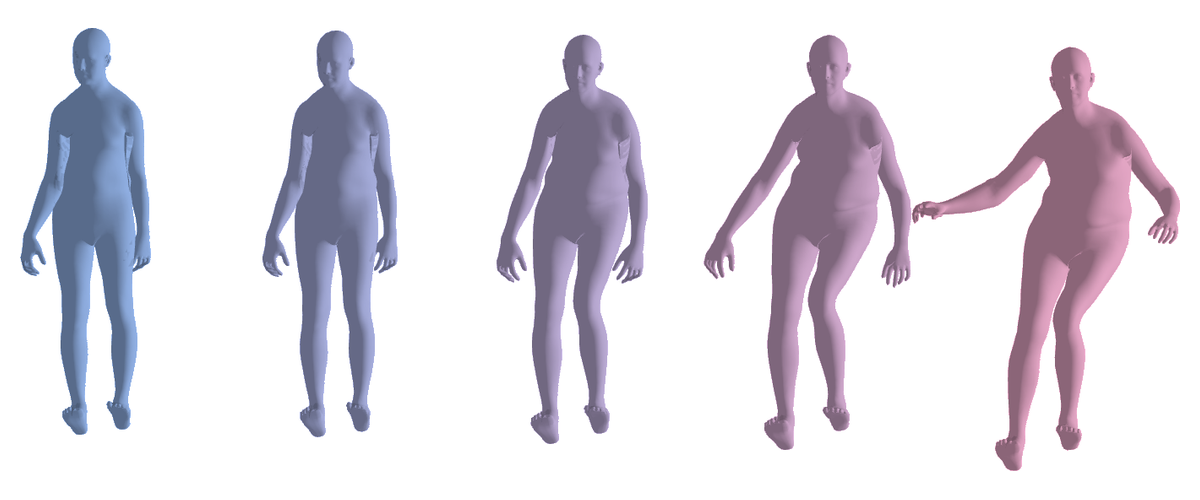}
        &
        \imgcell{\linewidth}{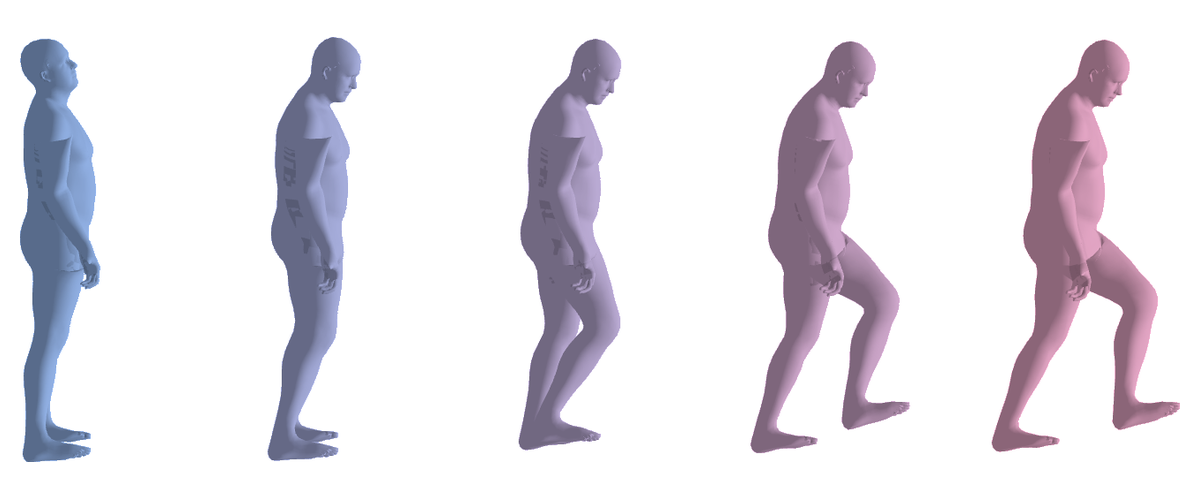}
        &
        \imgcell{\linewidth}{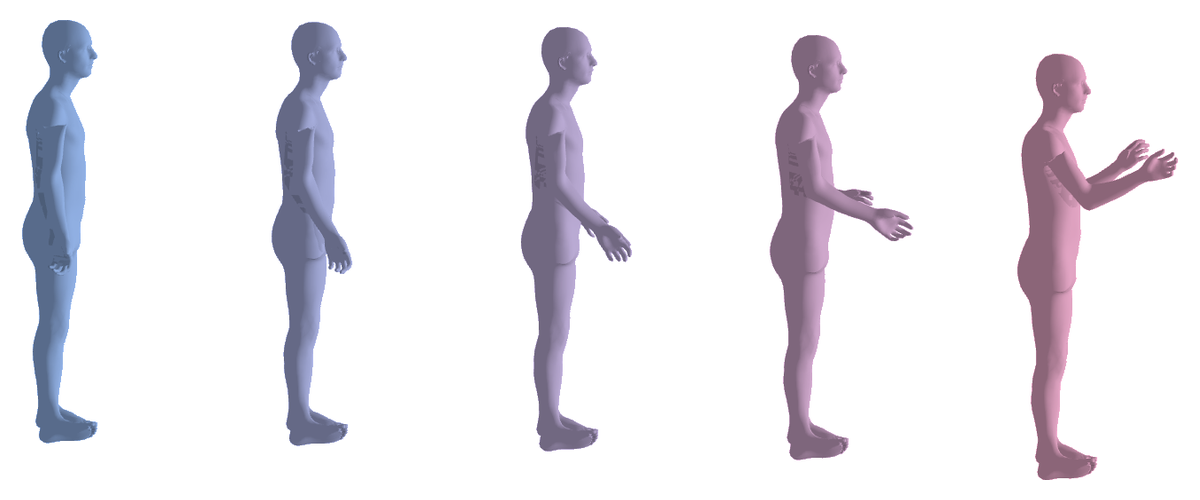}
        \\
        \midrule
        \lcell{Text\\Description}
        &
        \parbox{\linewidth}{\centering\footnotesize A man takes a step forward, takes his left arm and moves it right to left \bluetxt{then takes a step back}.}
        &
        \parbox{\linewidth}{\centering\footnotesize A person jumps \bluetxt{to his left}.}
        &
        \parbox{\linewidth}{\centering\footnotesize This person \bluetxt{stumbles left} and right while moving forward.}
        &
        \parbox{\linewidth}{\centering\footnotesize A person \bluetxt{claps their hands}.}
        \\
        \midrule
        \lcell{VQ-VAE Predicted Text}
        &
        \parbox{\linewidth}{\centering\footnotesize A person walking forward and then bending down to pick up something}
        &
        \parbox{\linewidth}{\centering\footnotesize a person jumps and lands.}
        &
        \parbox{\linewidth}{\centering\footnotesize a person walks forward and then walks sideways to the left.}
        &
        \parbox{\linewidth}{\centering\footnotesize a person juggles two balls with their hands.}
        \\
        \midrule
        \lcell{Ours}
         &
        \parbox{\linewidth}{\centering\footnotesize A person steps forward, picks something up with their left hand, \bluetxt{and then steps back}.}
        &
        \parbox{\linewidth}{\centering\footnotesize A person jumps sideways \bluetxt{to the left}.}
        &
        \parbox{\linewidth}{\centering\footnotesize A person walks forward and then \bluetxt{stumbles to the left}.}
        &
        \parbox{\linewidth}{\centering\footnotesize Person is \bluetxt{clapping their hands}.}
        \\
        \bottomrule

    \end{tabular}
    \label{tab:text-eval}
    \vspace{-2mm}
\end{table*}

\paragraph{Removing every proposed component.}
The \textit{Baseline} row of Table~\ref{tab:ablation} ablates all of our designs at once, using neither geometry-preserving initialization nor orthogonal regularization.
It reaches an Average of 45.77, while the full model reaches 53.48, a relative increase of \textbf{16.8\%}. Retrieval and captioning metrics both improve, although the gains vary by metric.
The same holds on KIT-ML, where the corresponding baseline reaches 36.52 against 41.71 (Table~\ref{tab:kit_ablation}).

\paragraph{Impact of Tokenizer Design.}
Replacing PQ-VAE~\cite{hong2025egolm} with DVQ improves caption quality and alignment together (CIDEr: $44.06 \rightarrow 59.71$, MMDist: $2.920 \rightarrow 2.680$).
R@1 is essentially unchanged ($0.534$ vs.\ $0.533$) while R@2 and R@3 both improve, and Average rises by \textbf{14.2\%} ($46.84 \rightarrow 53.48$).

\paragraph{Impact of Sparse Projection.}
Compared with learned linear projection, sparse projection improves MMDist from 4.20 to 2.68 and CIDEr from 33.54 to 59.71. Average increases by \textbf{34.5\%} ($39.77 \rightarrow 53.48$), showing that the initialization choice matters in this setting.

\paragraph{Impact of Regularization Method.}
Replacing pairwise cosine-distance regularization~\cite{wang2018cosface} with orthogonality regularization improves every metric reported in the HumanML3D ablation, including MMDist ($3.07 \rightarrow 2.68$) and CIDEr ($43.71 \rightarrow 59.71$). Average rises by \textbf{15.8\%} ($46.20 \rightarrow 53.48$).
The ordering also holds on KIT-ML ($38.32 \rightarrow 41.71$), though there the cosine variant keeps an edge on pure text-overlap metrics (Table~\ref{tab:kit_ablation}).

\paragraph{Impact of Sparse Projection-Based Initialization.}
Under the same orthogonal-loss ratio ($10^{-2}$), replacing sparse projection with stochastic initialization reduces Average by \textbf{8.3\%} ($53.48 \rightarrow 49.06$).
Figure~\ref{fig:total-loss-compare} is consistent with this trend: without sparse projection, training converges more slowly and remains at higher loss, indicating a less favorable optimization start.

\paragraph{Effect of Orthogonal Loss Ratio.}
Adding orthogonal regularization substantially improves over the no-orthogonal baseline, with the best ratio ($10^{-2}$) raising Average by \textbf{13.2\%}.
A stronger ratio ($10^{-1}$) gives the highest recalls but degrades captioning quality, and Figure~\ref{fig:total-loss-compare} shows faster, lower-loss optimization with the regularizer.
Across the tested range from $10^{-4}$ to $1$, Average stays between 49.50 and 53.48. The coefficient is therefore not highly sensitive within this single-run sweep, and $10^{-2}$ is also best among the tested KIT-ML settings.

\begin{table*}[t]
    \centering
    \small
    \setlength{\tabcolsep}{2.2pt}
    \caption{Ablation on KIT-ML under GPT-2, corresponding to Table~\ref{tab:ablation}. \textit{Baseline} removes all proposed components. Each tested replacement reduces the descriptive Average.}
    \renewcommand{\arraystretch}{1}
    \resizebox{\linewidth}{!}{
    \begin{tabular}{llcccccccccc}
    \toprule
    Study & Setting & R@1 & R@2 & R@3 & MMDist$\downarrow$ & BLEU-1$\uparrow$ & BLEU-4$\uparrow$ & R-L$\uparrow$ & CIDEr$\uparrow$ & BERTScore$\uparrow$ & Avg.$\uparrow$ \\
    \midrule
    Baseline & Init \ding{55}, R=0 & 0.140 & 0.269 & 0.380 & 6.048 & 49.38 & 19.85 & 44.02 & 45.56 & 34.00 & 36.52 \\
    \midrule
    Tokenizer & PQ-VAE & 0.121 & 0.214 & 0.280 & 7.893 & 45.42 & 15.91 & 43.25 & 37.98 & 34.85 & 32.99 \\
    Projection & Linear & 0.074 & 0.134 & 0.191 & 9.159 & 41.89 & 14.59 & 41.60 & 31.69 & 33.83 & 29.48 \\
    Regularization & Pairwise Cosine Dist. & 0.156 & 0.291 & 0.389 & 5.959 & \textbf{49.76} & \textbf{20.66} & \textbf{46.41} & 47.00 & \textbf{38.24} & 38.32 \\
    \midrule
    Ortho. Weight & Init \ding{51}, $R=0$ & 0.322 & 0.511 & 0.620 & 4.069 & 44.37 & 15.05 & 40.14 & 54.12 & 35.54 & 39.61 \\
    Ortho. Weight & Init \ding{51}, $R=10^{-2}$ (Ours) & \textbf{0.328} & \textbf{0.574} & \textbf{0.676} & \textbf{3.565} & 44.11 & 15.69 & 42.48 & \textbf{58.81} & 36.57 & \textbf{41.71} \\
    Ortho. Weight & Init \ding{51}, $R=1$ & 0.282 & 0.453 & 0.565 & 4.521 & 43.32 & 16.33 & 41.10 & 55.30 & 33.83 & 38.87 \\
    \bottomrule
    \end{tabular}
    }
    \label{tab:kit_ablation}
    \vspace{-3mm}
\end{table*}

\paragraph{Cross-dataset ablation on KIT-ML.}
Table~\ref{tab:kit_ablation} reproduces the component ordering on a smaller dataset. Replacing DVQ with PQ-VAE reduces Average from 41.71 to 32.99; learned projection gives 29.48, pairwise cosine regularization gives 38.32, and removing all components gives 36.52. The $10^{-2}$ orthogonal-loss ratio also outperforms both no regularization (39.61) and the over-strong ratio of 1 (38.87). The advantage is not uniform: pairwise cosine regularization remains better on the text-overlap metrics, whereas orthogonality is stronger on retrieval and CIDEr. We therefore treat KIT-ML as a directional cross-dataset check rather than an independent precision estimate.

\paragraph{Geometry and stability diagnostics.}
Two further checks directly measure the learned geometry and the stability of the discrete training path.

\begin{table}[t]
    \centering
    \small
    \setlength{\tabcolsep}{3pt}
    \caption{Geometry and stability diagnostics. \textbf{Top:} mean absolute pairwise cosine similarity (PCS); lower is closer to orthogonal. \textbf{Bottom:} removing the straight-through estimator (STE) from Stage-1 training.}
    \vspace{-2mm}
    \renewcommand{\arraystretch}{1}
    \begin{tabular*}{\linewidth}{@{\extracolsep{\fill}}lc@{}}
    \toprule
    Token space & Mean $|$PCS$|\downarrow$ \\
    \midrule
    VQ-VAE codebook & $9.62\times10^{-2}$ \\
    DVQ codebook (ours) & $\mathbf{4.51\times10^{-4}}$ \\
    GPT-2 text embeddings & $2.68\times10^{-1}$ \\
    Motion-token emb.\ (ours, tuned) & $\mathbf{2.24\times10^{-3}}$ \\
    \bottomrule
    \end{tabular*}

    \vspace{1.5mm}

    \begin{tabular*}{\linewidth}{@{\extracolsep{\fill}}lrr@{}}
    \toprule
    Stage-1 statistic & w/o STE & with STE \\
    \midrule
    Best total val.\ loss$\downarrow$ & 0.7849 & \textbf{0.5501} \\
    Final val.\ rec.\ loss$\downarrow$ & 0.9908 & \textbf{0.3589} \\
    Code utilization$\uparrow$ & 36.91\% & \textbf{100.00\%} \\
    Dead codes$\downarrow$ & 323/512 & \textbf{0/512} \\
    Max code count$\downarrow$ & 50,482 & \textbf{6,300} \\
    \bottomrule
    \end{tabular*}
    \label{tab:diag}
    \vspace{-4mm}
\end{table}

Table~\ref{tab:diag} (top) shows that the DVQ codebook is over two orders of magnitude closer to orthogonal than the VQ-VAE codebook ($4.51\times10^{-4}$ versus $9.62\times10^{-2}$). After sparse-projection initialization and Stage-2 fine-tuning, the motion-token embeddings retain a mean PCS of $2.24\times10^{-3}$. These values confirm approximate orthogonality within each space, but they do not measure one-to-one correspondence between tuned embeddings and their Stage-1 code vectors.

The lower panel isolates the STE while keeping the rest of the Gumbel-Softmax pipeline fixed. Without the STE, validation reconstruction loss initially falls to 0.2991 but rises to 0.9908 by the final epoch, and only 189 of 512 codes remain active. With the STE, all codes remain active, the final reconstruction loss is 0.3589, and the maximum code count falls from 50,482 to 6,300. The biased STE gradient is therefore important here because its hard forward path matches the discrete representation used at inference. Qualitative token sequences remain in Appendix~\ref{sec:appendix_case_studies}.

\vspace{-1mm}
\subsection{Case Study}
\label{sec:case_study}

Table~\ref{tab:text-eval} compares the VQ-VAE baseline and \sysname on representative motion sequences.
\sysname recovers the leftward direction in the second example and the clapping action in the fourth, which the VQ-VAE baseline misses. The comparison also exposes remaining errors: the first \sysname output adds a picking action that is absent from the reference, and the third captures only the leftward part of a left-and-right stumble. These examples illustrate behavior but do not establish a causal link between code usage and caption quality.
Token-level usage patterns and a second case study are provided in Appendix~\ref{sec:appendix_case_studies}.

\vspace{-2mm}
\section{Conclusion}
\vspace{-2mm}

We presented \sysname, which gives a motion codebook and its LLM motion-token embeddings a shared orthogonality prior. The method combines decoder-only DVQ, sparse isometric initialization, and regularization in both training stages. Under GPT-2, it raises a descriptive aggregate of retrieval and captioning metrics by \textbf{22.4\%} on HumanML3D and \textbf{14.4\%} on KIT-ML relative to the strongest reported baselines. Controlled ablations support the contribution of each component. Geometry diagnostics show low pairwise cosine similarity in the Stage-1 codebook and the tuned Stage-2 motion-token block; they do not establish direct semantic alignment with text embeddings.


\section*{Limitations}
Our ablations cover tokenizer design, projection strategy, regularization form, and orthogonal-loss ratio, but remain centered on our own components; we have not compared against other geometry-aware objectives such as whitening-, spectral-, or entropy-based constraints.

Our strongest results use fully fine-tuned GPT-2, while Qwen3 0.6B and Llama 3.2 1B are adapted with LoRA under our compute budget. The cross-backbone comparison therefore confounds model capacity with adaptation budget. Holding both fixed, the method improves Qwen3 0.6B (Section~\ref{sec:backbone_transfer}), but we have not tested full fine-tuning of larger instruction-tuned LLMs. All experiments use two English benchmarks with the same skeleton representation, and we evaluate understanding rather than motion generation. Because DVQ trades reconstruction loss for a better-conditioned codebook (Section~\ref{sec:codebook_analysis}), its effect on synthesis remains open.

Our geometric evidence is diagnostic rather than semantic. PCS measures pairwise orthogonality within a space; it does not establish semantic alignment or preservation after fine-tuning. Exact Gram preservation applies only at initialization and requires $K\leq D$; we test only $K=D=512$. Likewise, full utilization does not imply semantically distinct codes, and the qualitative examples do not replace blinded human evaluation.

The headline Average is a non-standard descriptive aggregate. Several published baselines omit R@2, so their recall term averages R@1 and R@3, whereas rows with complete reporting average R@1, R@2, and R@3. Individual metrics are therefore the primary evidence. We also report single values rather than means and confidence intervals for our main experiments, so the statistical uncertainty of the gains is unknown.




\section*{Acknowledgments}
We thank the anonymous reviewers for their constructive feedback, which
substantially improved the clarity and the empirical scope of this work.

\bibliography{citation}
\newpage
\appendix

\section{Appendix}


\begin{table*}[!b]
    \centering
    \caption{Token sequences for four ground-truth motions. DVQ uses a wider variety of token IDs in these examples; token IDs alone do not establish temporal coherence or semantic quality.}
    \vspace{-2mm}
    \setlength{\tabcolsep}{4pt}
    \renewcommand{\arraystretch}{1.0}

    \begin{tabular}{@{} m{0.13\textwidth} m{0.2\textwidth} m{0.2\textwidth} m{0.2\textwidth} m{0.2\textwidth} @{}}
        \toprule
        \lcell{GT Motions}
        & \imgcell{\linewidth}{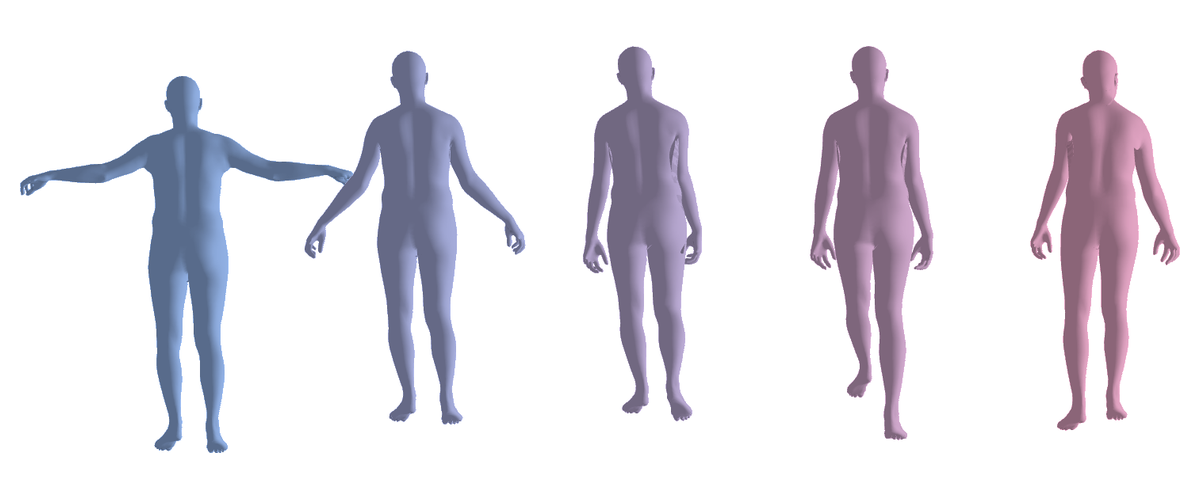}
        & \imgcell{\linewidth}{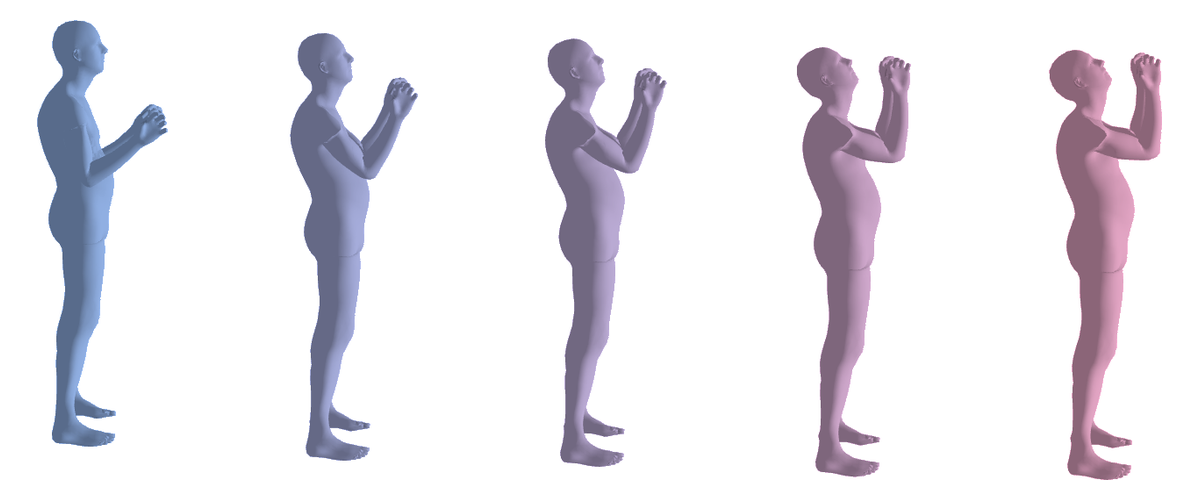}
        & \imgcell{\linewidth}{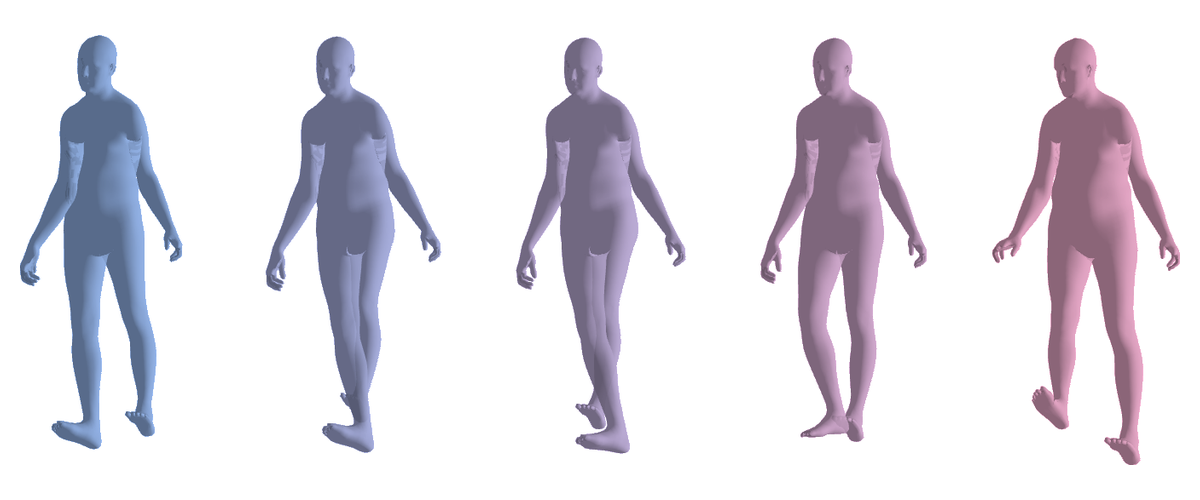}
        & \imgcell{\linewidth}{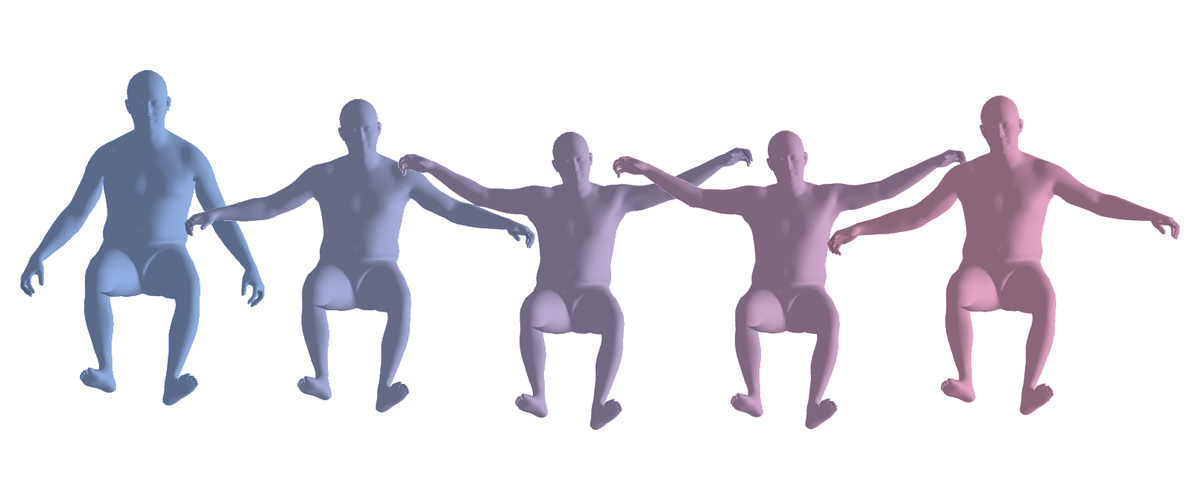}
        \\
        \midrule
        \lcell{Text\\Description}
        & \parbox{\linewidth}{\centering\footnotesize Person leans forward slightly and moves right hand in a wiping motion.}
        & \parbox{\linewidth}{\centering\footnotesize The person drinks from the big jug.}
        & \parbox{\linewidth}{\centering\footnotesize The man takes a step and bends raising a foot to wipe a table.}
        & \parbox{\linewidth}{\centering\footnotesize A person rests their hands on their knees while squatting.}
        \\
        \midrule
        \lcell{VQ-VAE Motion Token IDs}
        &
        \parbox{\linewidth}{\centering\footnotesize
        \tid{330} \tid{330} \tid{330} \tid{330} \tid{330} \tid{330} \tid{330} \tid{330}
        \tid{330} \tid{330} \tid{330} \tid{330} \tid{330} \tid{330} \tid{287} \tid{287}
        \tid{287} \tid{287} \tid{287} \tid{330} \tid{330} \tid{330} \tid{330} \tid{288}}
        &
        \parbox{\linewidth}{\centering\footnotesize
        \tid{493} \tid{243} \tid{243} \tid{243} \tid{243} \tid{243} \tid{243} \tid{243}
        \tid{243} \tid{243} \tid{248} \tid{28} \tid{10} \tid{10} \tid{10} \tid{173}
        \tid{173} \tid{280} \tid{119} \tid{128} \tid{153} \tid{153} \tid{153} \tid{153}}
        &
        \parbox{\linewidth}{\centering\footnotesize
        \tid{41} \tid{499} \tid{276} \tid{411} \tid{17} \tid{17} \tid{17} \tid{17}
        \tid{17} \tid{17} \tid{17} \tid{17} \tid{17} \tid{17} \tid{59} \tid{229}
        \tid{229} \tid{65} \tid{65} \tid{65} \tid{65} \tid{65} \tid{41}}
        &
        \parbox{\linewidth}{\centering\footnotesize
        \tid{296} \tid{296} \tid{296} \tid{27} \tid{27} \tid{27} \tid{27} \tid{27}
        \tid{27} \tid{27} \tid{27} \tid{27} \tid{27} \tid{27} \tid{27} \tid{27}
        \tid{27} \tid{27} \tid{27} \tid{27} \tid{27} \tid{27} \tid{27} \tid{27}}
        \\
        \midrule
        \lcell{DVQ Token IDs}
        &
        \parbox{\linewidth}{\centering\footnotesize
        \tid{379} \tid{19} \tid{379} \tid{177} \tid{177} \tid{343} \tid{189} \tid{225}
        \tid{385} \tid{330} \tid{343} \tid{343} \tid{177} \tid{343} \tid{177} \tid{19}
        \tid{352} \tid{330} \tid{414} \tid{385} \tid{12} \tid{385} \tid{385} \tid{414}}
        &
        \parbox{\linewidth}{\centering\footnotesize
        \tid{70} \tid{185} \tid{70} \tid{70} \tid{70} \tid{384} \tid{70} \tid{457}
        \tid{70} \tid{70} \tid{70} \tid{504} \tid{223} \tid{82} \tid{200} \tid{495}
        \tid{457} \tid{296} \tid{51} \tid{206} \tid{37} \tid{254} \tid{500} \tid{484}}
        &
        \parbox{\linewidth}{\centering\footnotesize
        \tid{337} \tid{232} \tid{156} \tid{345} \tid{370} \tid{19} \tid{370} \tid{370}
        \tid{19} \tid{370} \tid{370} \tid{370} \tid{370} \tid{38} \tid{38} \tid{358}
        \tid{421} \tid{284} \tid{219} \tid{500} \tid{72} \tid{55} \tid{219}}
        &
        \parbox{\linewidth}{\centering\footnotesize
        \tid{165} \tid{235} \tid{11} \tid{324} \tid{59} \tid{507} \tid{298} \tid{24}
        \tid{24} \tid{24} \tid{11} \tid{428} \tid{284} \tid{405} \tid{463} \tid{38}
        \tid{284} \tid{324} \tid{211} \tid{139} \tid{469} \tid{328} \tid{9} \tid{69}}
        \\
        \bottomrule
    \end{tabular}
    \label{tab:token_eval}
    \vspace{-2mm}
\end{table*}

\subsection{Additional Implementation Details}
\label{sec:appendix_dvq}

This section provides implementation details that complement the training protocol in Section~\ref{sec:experimental_settings}.

\paragraph{Backbone Adaptation.}
We evaluate three language backbones of increasing size: GPT-2, Qwen3 0.6B, and Llama 3.2 1B.
GPT-2 uses full-parameter fine-tuning: all transformer weights are updated together with the newly added motion-token embedding block, while the original text-token embeddings remain frozen as described in Section~\ref{sec:regulation}.
Qwen3 0.6B and Llama 3.2 1B are adapted with LoRA~\cite{hu2022lora} using rank 16 and scaling factor 32; the base weights stay frozen, and only the low-rank adapters and the motion-token embedding block receive gradient updates.
This split reflects our compute budget rather than a methodological requirement.
It also motivates the controlled comparison in Table~\ref{tab:qwen_controlled}, which holds the Qwen3 backbone and its LoRA configuration fixed and varies only our proposed components, so that the effect of the method is not confounded with the adaptation strategy.
For LLM fine-tuning, the cosine annealing scheduler uses $T_{\max}=200$ and $\eta_{\min}=10^{-6}$.

\paragraph{Data Preprocessing and Token Granularity.}
HumanML3D motions are sampled at 20 FPS and contain 20 to 200 frames after preprocessing; each frame is represented by the standard 263-dimensional feature vector, and each motion is paired with three text captions.
KIT-ML is processed with the same pipeline, using its 251-dimensional frame features, and provides a smaller, more diverse cross-dataset check.
Motion features are z-normalized with the training-set mean and standard deviation before entering the tokenizer.
DVQ operates on temporally downsampled motion features: three stride-2 convolutional stages shorten the sequence by a factor of 8 before quantization, so each discrete token summarizes a window of 8 frames, or 0.4 seconds at 20 FPS, rather than a single pose.
A maximum-length motion of 200 frames therefore maps to 25 tokens.
This reduces the length of the token sequences consumed by the language model, which lowers the fine-tuning cost and keeps motion-token sequences at a length scale comparable to their paired captions.

\paragraph{Tokenizer Architecture.}
The DVQ encoder is a one-dimensional convolutional network.
An input convolution lifts the frame features to 512 channels; three downsampling stages follow, each consisting of a stride-2 convolution and a residual block of depth 3 with dilation growth rate 3; a final convolution maps the 512 channels to $K=512$ logits per temporal position.
The encoder therefore emits the assignment logits $\mathbf{z}=Q(x)$ of Eq.~\ref{eq:gumbelsoftmax} directly, and token assignment involves no distance computation against the codebook.
The decoder mirrors the encoder with nearest-neighbor upsampling and convolutions, and its input consists exclusively of 512-dimensional code embeddings.
Stage-1 training uses randomly cropped 32-frame windows, corresponding to 4 tokens per training clip; full sequences are tokenized at inference.

\paragraph{DVQ Optimization.}
DVQ uses a 512-code, 512-dimensional codebook trained for 500 epochs with batch size 512 under AdamW.
The base learning rate is $2\times10^{-4}$ with 3\% linear warmup, and the cosine schedule described in the main text decays the learning rate to zero over the remaining epochs.
We apply weight decay of $10^{-4}$ to non-bias, non-normalization parameters and no weight decay to bias terms, normalization layers, or quantizer parameters.
Quantizer parameters, including the codebook, use the reduced learning rate of $10^{-4}$ (half the base rate), which keeps codebook updates conservative relative to the rest of the network while the assignment distribution is still changing.
The codebook is initialized from a Gaussian with standard deviation 0.02, matching the scale of standard transformer embedding initialization.
After training, we retain the checkpoint with the lowest total validation loss rather than the final-epoch weights, and this checkpoint is used both for reporting Stage-1 statistics and for tokenizing the corpus consumed in Stage~2.

\paragraph{Temperature and Utilization Schedules.}
Two schedules govern the discrete path during Stage-1 training.
First, the Gumbel-Softmax temperature $\tau$ stays at $0.4$ for the first 300 epochs, decays exponentially to $0.01$ over the next 100 epochs, and then remains fixed.
A moderate early temperature keeps the relaxed assignments informative enough for gradients to reach many codes, while the late low-temperature phase sharpens the relaxation toward the hard assignments used at inference.
Second, the utilization statistic $\mathbf{q}$ in Eq.~\ref{eq:loss_util} is computed from soft assignment probabilities for the first 150 epochs; a hardness coefficient then increases linearly from 0 to 1 over the following 50 epochs, after which the statistic is computed from hard assignments.
In the implementation, $\mathbf{q}$ is the mini-batch mean of the assignment probabilities over all temporal positions, and the objective is expressed through the normalized perplexity $\exp(H(\mathbf{q}))/K$, a monotone transform of the entropy in Eq.~\ref{eq:loss_util}; the hardness coefficient interpolates between the perplexity of the relaxed assignments and that of the straight-through hard assignments.
This soft-to-hard transition lets the utilization loss spread probability mass across the codebook early in training, and later aligns the regularized statistic with the actual discrete code usage.
Training uses stochastic Gumbel-Softmax samples throughout, whereas inference applies deterministic $\arg\max$ tokenization.

\paragraph{Regularization Coefficients.}
Unless otherwise stated, the Stage-1 objective uses $\lambda_{\mathrm{ortho}}=10^{-2}$ and $\lambda_{\mathrm{util}}=10^{-2}$, and the Stage-2 instruction-tuning objective uses $\lambda_{\mathrm{orth}}'=10^{-2}$, which is the setting denoted \textit{Init~\ding{51}, $R=10^{-2}$} in Table~\ref{tab:ablation}.
As reported in Section~\ref{sec:ablations}, downstream performance is stable across four orders of magnitude of this coefficient, with the descriptive Average staying between 49.50 and 53.48 over the tested range from $10^{-4}$ to $1$ on HumanML3D.
The coefficient therefore does not require careful per-dataset tuning, and the same value is used for both HumanML3D and KIT-ML.

\paragraph{Decoder-Only Design Rationale.}
DVQ's decoder exclusively consumes code embeddings, so no continuous latent path can bypass the codebook during reconstruction.
All information available to the decoder passes through the discrete code selection, so the reconstruction gradient must shape the code vectors themselves instead of flowing through a continuous shortcut around them.
Combined with the Gumbel-Softmax relaxation, this design allows the reconstruction, utilization, and orthogonality terms to be optimized jointly in a single computation graph, without the surrogate codebook and commitment losses used by standard VQ-VAE training.

\subsection{Gumbel-Softmax and Straight-Through Details}
\label{sec:appendix_gumbel}

For completeness, we provide the detailed definition of the Gumbel-Softmax operator~\cite{jang2017categorical} used in DVQ, which is omitted from the main text.
The $\mathrm{GumbelSoftmax}(\cdot; \cdot)$ used in Eq.~\ref{eq:gumbelsoftmax} is defined as:
\begin{equation}
\mathrm{GumbelSoftmax}(\mathbf{z}; \tau)
=
\mathrm{softmax}
\left(
\frac{\mathbf{z} + \mathbf{g}}{\tau}
\right),
\end{equation}
where $\mathbf{g}$ is sampled element-wise from the Gumbel$(0,1)$ distribution via $g_k = -\log(-\log u_k)$ with $u_k \sim \mathrm{Uniform}(0,1)$, and $\tau$ is the temperature parameter controlling the smoothness of the categorical distribution.
As $\tau \rightarrow 0$, the output approaches a one-hot vector, recovering discrete sampling from the categorical distribution defined by the logits; as $\tau$ grows, the output flattens toward a uniform mixture.

With the straight-through estimator~\cite{bengio2013estimating}, the forward pass discretizes the relaxed sample,
\begin{equation}
\mathbf{y}_{\text{hard}}
=
\mathrm{one\text{-}hot}
\left(
\arg\max_{k} \; y_{\text{soft},k}
\right),
\end{equation}
while the backward pass propagates gradients through $\mathbf{y}_{\text{soft}}$ as if the discretization were the identity map.
The resulting gradient is biased, but the forward computation matches the hard, discrete representation used at inference. In practice this property is decisive: as Table~\ref{tab:diag} reports, removing the straight-through estimator while keeping the rest of the pipeline fixed destabilizes late training (final validation reconstruction loss 0.9908 versus 0.3589) and collapses utilization, with 323 of 512 codes left unused.
Gumbel noise is injected only during training; at inference, tokenization is a deterministic $\arg\max$ over the logits, so the discrete motion-token sequence for a given input is unique and reproducible.

\subsection{Additional Discussion of the Sparse Projection}
\label{sec:appendix_projection}

Three properties of the sparse projection in Section~\ref{sec:projection} are worth stating explicitly.

First, it adds no learnable parameters.
The projection matrix $\mathbf{P}\in\{0,1\}^{D'\times D}$ is fully determined by the index set $\mathcal{I}$ of $D$ unique target rows; no weights are trained, and applying $\mathbf{P}$ amounts to scattering the $D$ code coordinates into $D$ fixed positions of a $D'$-dimensional zero vector.
Its memory and compute overhead are therefore negligible compared with a learned dense map.

Second, the guarantee is independent of the random choice of $\mathcal{I}$.
Any selection of $D$ unique rows yields orthonormal columns and hence $\mathbf{P}^\top\mathbf{P}=\mathbf{I}_D$, so the transferred Gram matrix, and with it the exact degree of approximate orthogonality attained in Stage~1, does not depend on the sampled indices.
The initialization is consequently insensitive to this source of randomness, in contrast to a learned linear projection whose outcome depends on its own optimization trajectory.
The ablation in Table~\ref{tab:ablation} quantifies this difference: replacing the sparse projection with a learned linear projection reduces the descriptive Average from 53.48 to 39.77 on HumanML3D, and the corresponding KIT-ML replacement is the weakest tested variant (Table~\ref{tab:kit_ablation}).

Third, the projection is an initialization mechanism only.
After Stage~2 fine-tuning begins, the motion-token embeddings are free parameters, and the equality between the codebook Gram matrix and the embedding Gram matrix holds only at the first step.
The Stage-2 orthogonality term then acts as a soft geometric prior rather than a constraint tying each embedding to its originating code; Table~\ref{tab:diag} shows that the tuned embeddings nevertheless retain a mean absolute pairwise cosine similarity of $2.24\times10^{-3}$, far below that of the surrounding GPT-2 text embeddings ($2.68\times10^{-1}$).

\subsection{Motion Vocabulary and Instruction Format}
\label{sec:appendix_vocab}

Stage~2 represents motions as text-like token strings.
The LLM vocabulary is extended with $K+3$ special tokens: \texttt{<motion\_id\_0>} through \texttt{<motion\_id\_511>} correspond to the DVQ codes, and three additional tokens mark the start of a motion span, the end of a motion span, and a mask placeholder reserved for masked-completion prompts.
For GPT-2, these tokens are appended after the original 50{,}257-entry text vocabulary, and their embeddings form the motion-token block $\mathbf{E}\in\mathbb{R}^{K\times D'}$ of Section~\ref{sec:regulation} with $D'=768$.
A tokenized motion is rendered as the start-of-motion token, the sequence of its code tokens, and the end-of-motion token, in that order.

The motion-to-text task uses a fixed instruction template: the input is \texttt{Generate text: <Motion\_Placeholder>}, where the placeholder is replaced by the rendered motion string, and the target output is one of the reference captions.
Training minimizes the autoregressive cross-entropy of Eq.~\ref{eq:tuning} on the target caption.
Consistent with Section~\ref{sec:regulation}, the original text embeddings stay frozen during fine-tuning; only the motion-token embedding block and the trainable LLM parameters receive gradients.

\subsection{Evaluation Details and Compute}
\label{sec:appendix_eval}

We follow the motion-to-text evaluation protocol of MotionGPT3~\cite{motiongpt3}.
The retrieval metrics (R-Precision at ranks 1--3 and MMDist) assess how well the generated caption aligns with its source motion in a shared text--motion embedding space, whereas the captioning metrics (BLEU-1, BLEU-4~\cite{bleu}, ROUGE-L~\cite{rouge}, CIDEr~\cite{cider}, and BERTScore~\cite{bertscore}) compare the generated caption against the reference captions, of which HumanML3D provides three per motion.
The two families probe different failure modes: a caption can be lexically close to a reference while retrieving poorly, and vice versa, which is why Section~\ref{sec:motion_understanding} reports both and treats the aggregate Average as descriptive only.
As noted in the main text, several published baselines omit R@2, so their recall term averages R@1 and R@3 while fully reported rows average all three cutoffs; this asymmetry is a further reason not to read the aggregate as a benchmark measure.

Unless otherwise specified, each experiment runs on a single NVIDIA B200 GPU.
Stage-1 DVQ training uses 500 epochs at batch size 512, and Stage-2 motion understanding uses 100 epochs at batch size 320 with learning rate $10^{-4}$, weight decay $10^{-2}$, and cosine scheduling.
We report single-run results without confidence intervals, a limitation discussed in the Limitations section.


\subsection{Qualitative Case Study on Codebook Usage}
\label{sec:appendix_case_studies}

Table~\ref{tab:token_eval} lists the discrete token sequences produced by the VQ-VAE baseline and by DVQ for four ground-truth motions, alongside their reference captions.
The examples were chosen to span different motion types: a localized upper-body action (wiping), an object-directed action (drinking), a compound locomotion-and-manipulation sequence (stepping and wiping a table), and a largely static pose (squatting with hands on knees).

Two patterns are visible.
First, the VQ-VAE baseline tends to assign long runs of a single code.
In the first example, code 330 covers 18 of the 24 positions and the entire sequence uses only three distinct codes; in the fourth example, a single code (ID 27) accounts for 21 of 24 positions.
DVQ tokenizes the same motions with a wider variety of code IDs, using 11 and 19 distinct codes on these two sequences, respectively.
Second, the contrast is smallest where the motion itself is varied: on the drinking and table-wiping examples, the VQ-VAE sequences already switch codes several times, and DVQ mainly redistributes usage rather than adding diversity.

These per-example observations are consistent with the aggregate usage statistics in Table~\ref{tab:codebook_stats}, where DVQ shows a lower maximum code count, standard deviation, coefficient of variation, and Gini coefficient than the VQ-VAE baseline under the identical codebook size.

We emphasize the limits of this evidence.
Token IDs alone do not establish temporal coherence or semantic quality: a long run of one code can be an appropriate encoding for a near-static segment, such as the held squat in the fourth example, and a diverse sequence is not automatically a better one.
The concern raised by the baseline is therefore not repetition per se, but the concentration of assignments onto few codes across many different motions, which the dispersion statistics of Section~\ref{sec:codebook_analysis} measure directly.
We accordingly treat these sequences as illustrations of the tokenizers' qualitative behavior, and rely on the aggregate distribution statistics and the downstream understanding results of Section~\ref{sec:motion_understanding} for quantitative conclusions.

\end{document}